\documentclass[10pt, a4paper]{article}
\usepackage{lrec2014}
\usepackage{graphicx}

\usepackage{epstopdf}
\usepackage[latin1]{inputenc}
\usepackage{multirow}
\usepackage{array}
\newcolumntype{C}[1]{>{\centering\arraybackslash}p{#1}}

\title{Language Models of Spoken Dutch\thanks{Support is acknowledged from IWT--INNOVATIEF AAN\-BESTEDEN and VRT in the STON project.}}

\name{Lyan Verwimp$^1$, Joris Pelemans$^1$, Marieke Lycke$^2$, Hugo Van hamme$^1$, Patrick Wambacq$^1$}

\address{ Affiliation1, Affiliation2, Affiliation3 \\
               Address1, Address2, Address3 \\
               author1@xxx.yy, author2@zzz.edu, author3@hhh.com\\}

\address{$^1$Dept. ESAT, Katholieke Universiteit Leuven, Belgium \\
$^2$VRT, Belgium \\
\small \tt {\{lyan.verwimp,joris.pelemans,hugo.vanhamme,patrick.wambacq\}}@esat.kuleuven.be\\
\small \tt marieke.lycke@vrt.be\\}

\begin{document}

\maketitleabstract

\section{Introduction}
\label{sec:intro}


In Flanders, all TV shows are subtitled. However, the process of subtitling is a very time-consuming one and can be sped up by providing the output of a speech recognizer run on the audio of the TV show, prior to the subtitling. Naturally, this speech recognition will perform much better if the employed language model is adapted to the register and the topic of the program. 

We present several language models trained on subtitles of television shows provided by the Flemish public-service broadcaster \textit{VRT}. This data was gathered in the context of the project \textit{STON} which has as purpose to facilitate the process of subtitling TV shows. One model is trained on all available data (46M word tokens), but we also trained models on a specific type of TV show or domain/topic. Language models of spoken language are quite rare due to the lack of training data. The size of this corpus is relatively large for a corpus of spoken language (compare with e.g.\ \textit{CGN} which has 9M words), but still rather small for a language model. Thus, in practice it is advised to interpolate these models with a large background language model trained on written language. The models can be freely downloaded on http://www.esat.kuleuven.be/psi/spraak/downloads/.




\section{Data}
\label{sec:data}

The data on which the models are trained consists of several types of television programs: documentaries, fiction, talkshows, daily news, weather reports, one quiz and one lifestyle show. In total we have 57 different TV shows. In table \ref{tab:types}, the number of episodes and number of word tokens for each type of show are displayed. A language model trained on all data available is released, together with a smaller model for each type of TV show, as shown in table \ref{tab:types}.

\begin{table}[b]
\centering
\begin{tabular}{|l|c|c|}
\cline{1-3}
			&\# of episodes&\# of word tokens \\
\hline
fiction		&3760&13.4M \\
\hline
documentary &3561&12.5M \\
\hline
talkshow	&1318&11.7M \\
\hline
quiz 		&2047&8M \\
\hline
lifestyle	&258&472k \\
\hline
news/weather &13&52k\\
\hline
\textbf{total}		&10957&46M \\
\hline
\end{tabular}
\caption{Number of episodes and number of word tokens for each \textbf{type} of television show.}
\label{tab:types}
\end{table}


Firstly, fiction is the largest group of television shows. The majority of them are adult programs (11M words): 2 crime fiction shows (142 episodes), 1 other fiction show (6 episodes) and 1 daily soap, of which we have subtitles of 2558 episodes, yielding 10M word tokens. Besides the adult shows, the data set also contains 7 children's programs (1046 episodes or 2.4M word tokens).


Secondly, the documentaries can be divided into two main groups: documentaries with only voice-over and documentaries with interviews. The first group consists of 25 nature movies, delivering 87k word tokens. The voice-over typically closely follows the screenplay, thus the language is not very spontaneous. The second type, documentaries with interviews, is a very heterogeneous group in terms of topic. Several programs even belong to different topics at the same time, such as \textit{God in Frankrijk}, a documentary about the Tour de France (sports) and France in general (traveling). Since there are interviews, the language in these documentaries is more spontaneous and dialectal. We have trained language models on subtitles of 35 different programs, in total 3536 episodes (for some programs we have only a few episodes, for others several hundreds) or 12.4M word tokens.

Our data set also contains the subtitles of 6 talkshows: 2 of them are about recent events and topics (291 episodes or 2.5M words), 2 about political and social themes (395 episodes or 4.6M words) and 2 about soccer (632 episodes or 4.5M words). 

The subtitles of the news report and the weather report together contain 52k word tokens. Finally, we have 2047 episodes of a daily quiz, good for approximately 8M words, and 258 episodes of a lifestyle program about the everyday life of ordinary people, with many interviews (472k words).

Since the data comprises different domains and since different types of TV shows can be about the same domain (e.g.\ there are sports documentaries but also talkshows about sports), we also trained several smaller language models focused on a single domain. The different domains can be found in table \ref{tab:domain}, along with the number of shows belonging to the domain and the total number of word tokens. We have to note that the domain ``sports" consists largely of soccer-related TV shows, next to only a few episodes about bob-sleighing, cycling and sports people in general. The domains of ``human interest" and ``current topics" cover a very diverse series of topics. 


\begin{table}
\centering
\begin{tabular}{|l|c|c|}
\cline{1-3}
			&\# of shows&\# of word tokens \\
\hline
general fiction	&2&10M \\
\hline
current topics	&8&9.5M \\
\hline
politics/society	&5&5.2M \\
\hline
sports		&5&4.6M \\
\hline
human interest 		&18&2.9M \\
\hline
children		&7&2.5M \\
\hline
traveling		&6&2.4M \\
\hline
police/justice 		&3&860k \\
\hline
nature &3&460k \\
\hline
history &3&380k\\
\hline
medical			&2&280k \\
\hline
love			&2&139k \\
\hline
music	&1&69k \\
\hline
\end{tabular}
\caption{Number of television shows and number of word tokens for each \textbf{domain/topic}. Several shows belong to two domains. All shows except the quiz are included in this classification.}
\label{tab:domain}
\end{table}

\section{Preprocessing}
\label{sec:norm}

In this section, we discuss the preprocessing that was necessary in order to obtain text suitable for language model training. This preprocessing is based on the one described in \cite{Demuynck2009}. The purpose of this process is to obtain the correct (and consistent) format for all training and test corpora, but it has also other advantages which will be discussed below. The preprocessing  consists of three major steps: normalization, uppercase conversion and spelling correction.

During the first step, the normalization, several symbols and measures are written in full such that more appropriate pronunciations can be generated for the lexicon, and metadata is removed. As regards punctuation, the end-of-sentence punctuation is used to split lines that contain more than one sentence and to merge sentences that are spread over different lines. After the splitting and merging process, all punctuation is removed. For dots and apostrophes, it is first checked whether they are part of an abbreviation or contraction; if that is the case, the abbreviated form is written in full. All numerical items are written in full and split (e.g.\ 274 becomes \textit{twee honderd vier-en zeventig} rather than \textit{tweehonderdvierenzeventig}). Splitting the numbers helps to reduce the storage space and helps generalizing to unseen numbers. Lines that only contain capital words are removed, because they typically contain script information and not spoken utterances. If the line contains a mixture of uppercase and lowercase words, the uppercase words are removed if they are longer than 4 letters: if the word is 4 letters or shorter, it is more likely to be an abbreviation or acronym than script information. Finally, trailing spaces are removed and begin- and end-of-sentence tokens are added.


In the second step of the preprocessing, sentence-initial words are converted to lowercase if their frequency is lower than their lowercase variant in a frequency list of lower- and uppercase words, as this is an indication that the word is only capitalized because of its sentence-initial position.


The last step of the preprocessing corrects spelling errors (e.g.\ \textit{on-line} $\rightarrow$ \textit{online}) and maps different orthographic variants to a single canonical form to ensure consistency (e.g.\ \textit{Schelde-oever}~$\rightarrow$~\textit{Scheldeoever} ``shore of the Scheldt").



\section{Models}
\label{sec:model}

The language models were trained with \textit{SRILM} \cite{Stolcke2002}. They are all open-vocabulary 5-gram models with modified Kneser-Ney smoothing \cite{ChenGoodman1999} and no count cut-offs. We trained one model on all the data, 6 language models on each type of TV show (see the rows in table \ref{tab:types}) and 13 language models on each domain (see the rows in table \ref{tab:domain}). The models are released both as count files, such that it is possible to train other language models than the ones provided, and as language models that can readily be used (in \textit{ARPA} format).




\section{Speech recognition}
\label{sec:exp}

\subsection{Set-up}
\label{sec:setup}

The speech recognition experiments were done using the \textit{SPRAAK} toolkit \cite{Demuynck2008}, configured according to \cite{Demuynck2009}, although the preprocessing is slightly different. The acoustic model for this recognizer was trained on broadcast news. We compare models trained on \textit{Mediargus}, a collection of 22 newspapers in Dutch (1.2B words); components $a$, $b$, $c$, $d$, $e$, $f$, $i$, $j$, $k$, $l$ and $m$ of \textit{CGN} (250k words); all data of \textit{VRT} (``\textit{VRT} all");  1 model trained on a specific type of TV show (documentary ``docu") and 2 models trained on a specific domain (general fiction ``gen-fic" and current topics ``current" ). The vocabularies contain all the words in the training text, except for the model trained on \textit{Mediargus} (limited to 400k).


We test the language models on several test sets: the first test set is a part of component $g$ (henceforth referred to as ``comp-g") of the \textit{Corpus of Spoken Dutch} (\textit{CGN}) \cite{Oostdijk2000}, which contains recordings of discussions, debates and meetings (25k word tokens or 2.88h of audio). The other test sets consist of television programs provided by \textit{VRT}: an episode of a daily soap (``soap", 30min or 8k word tokens) and a documentary with interviews about current topics (``docu-i", 53min or 10k words).

\subsection{Speech recognition results}
\label{sec:asr}

Table \ref{tab:asr1} shows the results for speech recognition with language models trained on a single data set. The model of \textit{Mediargus} performs the best for \textit{comp-g} and \textit{docu-i}, which is not surprising given the fact that it is trained on much more data than the other two models. Nevertheless, for the soap -- which has very spontaneous and dialectal language (hence the very high word error rates) -- the language model trained on data of \textit{VRT} gives the best performance, although it is trained on a corpus that is ca. 26 times smaller than the corpus of \textit{Mediargus}. 

\begin{table}
\centering
\begin{tabular}{|l|c|c|c|}
\cline{2-4}
\multicolumn{1}{c|}{}&\multicolumn{3}{c|}{test set} \\
\hline
model			&comp-g&soap&docu-i \\
\hline
\textit{Mediargus}  	&\textbf{28.2}&79.7&\textbf{38.1} \\
\hline
\textit{CGN}			&35.1&76.8&42.8 \\
\hline
\textit{VRT} all		&30.8&\textbf{75.0}&39.1 \\
\hline
\end{tabular}
\caption{Word error rates for 3 test sets, for models trained on a single training set.}
\label{tab:asr1}
\end{table}


In table \ref{tab:asr2}, results for the interpolation of \textit{Mediargus}, \textit{CGN} and the large model of \textit{VRT} are shown, where the interpolation weights are calculated on respectively another part of component $g$ of \textit{CGN} (\textit{comp-g-dev}), a set of subtitles from the soap (\textit{soap-dev}) and from a documentary with interviews (\textit{docu-i-dev}). The underlined WERs mark the best results overall for a certain test set. The interpolation of all data available optimized on \textit{comp-g-dev} obtained the best results for \textit{comp-g} (WER reduction of 6.7\% relative). Table \ref{tab:asr3} shows the results for an interpolation of \textit{Mediargus}, \textit{CGN} and a type or domain model. We see that using a specific model did not improve the recognition of \textit{comp-g}, since none of the in-domain language models we tested was relevant for this test set.

If we look at the results for the second test set, we see that the soap benefits the most from the use of language models of spoken language. Both in the interpolation with all data of \textit{VRT} and in the interpolation with only the fiction data, the \textit{VRT} language model has a high weight (0.83 and 0.82 respectively). Given that we have subtitles of 2558 episodes for this soap, which results in a language model trained on 10M words, this is not surprising. The best results were obtained by only using fiction data (see table \ref{tab:asr3}) (3\% relative WER reduction with respect to the language model of \textit{VRT} data only and 8\% relative WER reduction with respect to the model of \textit{Mediargus}). Nevertheless, the recognition is still poor, probably due the fact that the acoustic model and pronunciation lexicon are not adapted to the dialectal speech.

The third test is a documentary with interviews about a current topic. We see that adding all the \textit{VRT} data produces the best results (table \ref{tab:asr2}) (8.9\% relative WER reduction with respect to \textit{Mediargus} only). Optimizing the interpolation weights on \textit{comp-g-dev} worked better than optimizing them on \textit{docu-i-dev}, but since both the test set and the development set for the documentary consist of one episode only, we cannot draw any conclusions about this unexpected result. Only adding documentaries or TV shows about current topics does not improve (table \ref{tab:asr3}) with respect to using the large \textit{VRT} language model (but it does improve with respect to using \textit{Mediargus} only, see table \ref{tab:asr1}), probably because the theme of this type of documentary changes for every new episode.

\begin{table}
\centering
\begin{tabular}{|l|c|c|c|}
\cline{2-4}
\multicolumn{1}{c|}{}&\multicolumn{3}{c|}{test set} \\
\hline
optimized on			&comp-g&soap&docu-i \\
\hline
comp-g-dev			&\underline{\textbf{26.3}}&76.4&\underline{\textbf{34.7}} \\
\hline
soap-dev			&27.9&75.2&36.7 \\
\hline
docu-i-dev			&27.1&\textbf{74.5}&36.3 \\
\hline
\end{tabular}
\caption{Word error rates for 3 test sets for the interpolation of \textit{Mediargus}, \textit{CGN} and all data of \textit{VRT}, where the first column indicates the data set on which the interpolation weights were calculated.}
\label{tab:asr2}
\end{table}

\begin{table}
\centering
\begin{tabular}{|l|l|c|c|c|}
\cline{3-5}
\multicolumn{2}{c|}{}&\multicolumn{3}{c|}{test set} \\
\hline
optim. on&model		&comp-g&soap&docu-i \\
\hline
soap-dev&gen-fic 		&39.1&\underline{\textbf{72.7}}&43.6 \\
\hline
\multirow{2}{*}{docu-i-dev}&docu			&\textbf{28.8}&77.8&36.8 \\
\cline{2-5}
&current		&\textbf{28.8}&78.0&\textbf{35.9} \\
\hline
\end{tabular}
\caption{Word error rates for 3 test sets for the interpolation of \textit{Mediargus}, \textit{CGN} and a type or domain model, where the first column indicates the data set on which the interpolation weights were calculated 
and the second column the training set (type or domain). }
\label{tab:asr3}
\end{table}

\section{Conclusion}
\label{sec:concl}

We presented several language models of spoken Dutch (one large one and several small in-domain ones), trained on normalized subtitles of TV shows. Models of spoken language are quite rare and a valuable source for speech recognition, as our experiments with an interpolation of a large background model (of written language) with the smaller models of spoken language show.




\bibliographystyle{lrec2014}
\bibliography{LMspokenDutch}

\end{document}